%% file: main.tex
\documentclass[conference]{IEEEtran}
\IEEEoverridecommandlockouts

\usepackage{cite}
\usepackage{amsmath,amssymb,amsfonts}
\usepackage{algorithmic}
\usepackage{graphicx}
\usepackage{textcomp}
\usepackage{xcolor}
\usepackage{booktabs}
\usepackage{multirow}
\usepackage{url}

\def\BibTeX{{\rm B\kern-.05em{\sc i\kern-.025em b}\kern-.08em
    T\kern-.1667em\lower.7ex\hbox{E}\kern-.125emX}}

\begin{document}

\title{Automatic End-to-End Data Integration using Large Language Models}

\author{\IEEEauthorblockN{Aaron Steiner}
\IEEEauthorblockA{\textit{Data and Web Science Group} \\
\textit{University of Mannheim}\\
Mannheim, Germany \\
aaron.steiner@uni-mannheim.de}
\and
\IEEEauthorblockN{Christian Bizer}
\IEEEauthorblockA{\textit{Data and Web Science Group} \\
\textit{University of Mannheim}\\
Mannheim, Germany \\
christian.bizer@uni-mannheim.de}
}

\maketitle

\begin{abstract}
Designing data integration pipelines typically requires substantial manual effort from data engineers to configure pipeline components and label training data. While LLMs have shown promise in handling individual steps of the integration process, their potential to replace all human input across end-to-end data integration pipelines has not been investigated. As a step toward exploring this potential, we present an automatic data integration pipeline that uses GPT-5.2 to generate all artifacts required to adapt the pipeline to specific use cases. These artifacts are schema mappings, value mappings for data normalization, training data for entity matching, and validation data for selecting conflict resolution heuristics in data fusion. We compare the performance of this LLM-based pipeline to the performance of human-designed pipelines along three case studies requiring the integration of video game, music, and company related data. Our experiments show that the LLM-based pipeline is able to produce similar results, for some tasks even better results, as the human-designed pipelines. End-to-end, the human and the LLM pipelines produce integrated datasets of comparable size and density. Having the LLM configure the pipelines costs approximately \$10 per case study, which represents only a small fraction of the cost of having human data engineers perform the same tasks. 
\end{abstract}

\begin{IEEEkeywords}
data integration, large language models, schema matching, value normalization, entity resolution, data fusion, ETL pipelines
\end{IEEEkeywords}

\input{sections/01_introduction}
\input{sections/02_case_studies}
\input{sections/03_schema_matching}

\input{sections/04_normalization}

\input{sections/05_entity_matching}
\input{sections/06_data_fusion}

\input{sections/07_end_to_end}
\input{sections/08_related_work}
\input{sections/09_conclusion}

\bibliographystyle{IEEEtran}
\bibliography{main}

\end{document}

%% file: sections/01_introduction.tex
\section{Introduction}
\label{sec:introduction}
Data integration is the process of combining data from multiple heterogeneous data sources into a unified, coherent dataset~\cite{doan2012principles,naumann2006dataFusionThreeSteps,khatiwada2022integrating}. This process typically involves multiple steps: schema matching to align attributes across sources, value normalization to standardize data formats, entity matching to identify records referring to the same real-world entity, and data fusion to resolve data conflicts and create a single consolidated record for each entity~\cite{dong2009dataFusion,naumann2006dataFusionThreeSteps}. Traditionally, each step requires substantial manual effort from data engineers who must configure pipeline components and label training data.
Large language models (LLMs) have shown promise in reducing the manual effort required for individual data integration tasks~\cite{freire2025llmDataDiscoveryIntegration}: For schema matching, LLMs can leverage their background knowledge to capture attribute semantics to identify correspondences between heterogeneous schemas~\cite{narayan2022can, zhang2023schema,liu2025magneto}. For entity matching, LLMs have demonstrated competitive performance with BERT-based methods while requiring fewer labeled examples~\cite{peeters2024entity}. These advances raise the question of whether LLMs can successfully generate all artifacts that are required to adapt data integration pipelines to specific use cases, and thus automate the complete data integration process.

This paper investigates to what extent LLMs can replace human involvement in configuring end-to-end data integration pipelines. For this, we present an automatic end-to-end data integration pipeline that relies on GPT-5.2 to perform schema matching, generate value mappings for data normalization, generate training data for entity matching, and generate validation data for selecting conflict resolution heuristics for data fusion. We compare the performance of the automatic pipeline to human-designed pipelines. The human-designed pipelines were created by graduate students as part of a master course on data integration. The students manually labeled training and validation data and configured the pipeline components using knowledge from the course and trial-and-error experimentation. Both the LLM-based and the human-designed pipelines employ methods from the PyDI data integration framework\footnote{\url{https://github.com/wbsg-uni-mannheim/PyDI}}, an open-source library that implements traditional string-based as well as modern embedding- and LLM-based methods for schema matching, value normalization, entity matching, and data fusion. By relying on PyDI, both pipelines can choose scalable methods for tasks that require heavy data processing. The key difference lies in how each pipeline is configured, i.e., which methods are selected and how training and validation data is labeled.

We evaluate the pipelines through three case studies requiring the integration of heterogeneous data describing video games, companies, and music releases. Each use case requires the integration of three datasets with heterogeneous schemas, value formats, densities, and levels of data quality.

\input{tables/dataset_statistics}

The contributions of the paper are as follows:

\begin{enumerate}
    \item We present a fully automated end-to-end data integration pipeline that uses an LLM to perform all tasks that traditionally require human expertise, such as choosing methods, labeling training data, and generating validation sets. The pipeline covers the integration steps schema matching, value normalization, entity matching, and data fusion.

    \item We evaluate the automated pipeline along three data integration use cases and compare its results (step-wise and end-to-end) to the performance of human-configured pipelines established by graduate-level data engineers.

    \item We identify limitations of LLM-based methods that we employ and discuss how these limitations could potentially be addressed.

    \item We publish all artifacts of the three case studies (datasets, training, validation and test sets), contributing to the hopefully growing pool of evaluation data for end-to-end data integration pipelines. 
\end{enumerate}

The remainder of the paper is organized as follows. Section~\ref{sec:case_studies} describes the three case studies. Sections~\ref{sec:schema_matching} through~\ref{sec:data_fusion} describe how the LLM-based pipeline handles the different integration steps and compares its results to the human-configured pipelines. Section~\ref{sec:end_to_end} presents the end-to-end evaluation. Section~\ref{sec:conclusion} concludes with a discussion of findings and future work. All code and data for reproducing the experiments are available in the accompanying GitHub repository.\footnote{\url{https://github.com/wbsg-uni-mannheim/automatic-data-integration}}

%% file: tables/dataset_statistics.tex

\begin{table*}[t]
\caption{Dataset statistics for the three case studies.}
\label{tab:dataset_statistics}
\centering
\small
\begin{tabular}{llrrcp{9.5cm}}
\toprule
\textbf{Case Study} & \textbf{Source} & \textbf{Rows} & \textbf{Attrs} & \textbf{Density} & \textbf{Attributes} \\
\midrule
\multirow{4}{*}{\textbf{Games}}
 & DBpedia & 46,580 & 6 & 91.0\% & wiki\_ref, title, launch\_yr, studio, system, franchise \\
 & Metacritic & 20,494 & 8 & 97.7\% & mc\_id, game\_title, year\_published, made\_by, console, press\_rating, player\_rating, age\_rating \\
 & Sales & 7,877 & 10 & 98.7\% & rec\_id, prod\_title, launch\_dt, studio, dist, hw, press\_score, comm\_rating, age\_classification, units\_sold\_mm \\
\cmidrule{2-6}
 & \textit{Target Schema} & -- & 12 & -- & id, name, releaseYear, developer, genres, publisher, platform, criticScore, userScore, ESRB, globalSales, series \\
\midrule
\multirow{4}{*}{\textbf{Companies}}
 & Forbes & 2,000 & 7 & 99.2\% & forbes\_url, company, url, region, business\_segment, asset\_value, sales\_figure \\
 & DBpedia & 10,085 & 9 & 66.4\% & entity\_uri, org\_name, established, nation, headquarters, sector, keypeople\_name, total\_assets\_val, annual\_income \\
 & FullContact & 1,931 & 6 & 68.3\% & Attribute\_1, Attribute\_2, Attribute\_3, Attribute\_4, Attribute\_5, Attribute\_6 \\
\cmidrule{2-6}
 & \textit{Target Schema} & -- & 10 & -- & id, name, website, founded, country, city, industry, assets, revenue, founders \\
\midrule
\multirow{4}{*}{\textbf{Music}}
 & Discogs & 22,627 & 8 & 98.4\% & rec\_uid, title\_str, performer, pub\_dt, origin\_loc, imprint, category, tracks\_track-name \\
 & Last.fm & 9,865 & 5 & 78.7\% & item\_code, album\_title, band, tracks\_track-name, album\_length\_min \\
 & MusicBrainz & 4,763 & 7 & 83.3\% & Attribute\_1, Attribute\_2, Attribute\_3, Attribute\_4, Attribute\_5, Attribute\_6, Attribute\_7 \\
\cmidrule{2-6}
 & \textit{Target Schema} & -- & 8 & -- & id, name, artist, release-date, release-country, label, genre, tracks, duration \\
\bottomrule
\end{tabular}
\end{table*}

%% file: sections/02_case_studies.tex
\section{Case Studies}
\label{sec:case_studies}
We evaluate the pipelines using three case studies covering different domains. Each involves integrating three heterogeneous datasets into a unified dataset conforming to a predefined target schema. The target schemata were manually created.  Table~\ref{tab:dataset_statistics} provides statistics about the datasets including record and attribute counts, and data density. The table also lists the attributes of the datasets and the target schemata.

\textbf{Video Games.} \label{sec:games_case}
This case study requires the integration of data describing video games from three complementary sources: One dataset has been scraped from Metacritic and consists of review scores and ESRB ratings from the review aggregation platform; the second dataset is taken from Zenodo and contains global and regional sales figures. The third dataset was created by querying DBpedia and contains encyclopedic attributes such as release dates, developers, and series information.  The target schema of the games case study contains 12 attributes. Key challenges of this case study include matching games that appear on different platforms as separate entities, as well as normalizing platform and genre taxonomies across sources.

\textbf{Companies.} \label{sec:companies_case}
This case study integrates data describing large companies from three sources with distinct perspectives: the Forbes Global 2000 list ranks the world's largest public companies by market value, sales, profits, and assets; a DBpedia extract containing encyclopedic information including founding dates, headquarters, industry, and key people; and FullContact, a commercial API providing contact information and key personnel for large companies. The target schema of the companies use case contains 10 attributes including a taxonomic industry attribute. Key challenges include matching companies despite name variations (abbreviations, legal suffixes) and normalizing financial figures reported using different scales.

\textbf{Music.} \label{sec:music_case}
This case study requires the integration of data about music releases (albums, EPs, singles) from three community-maintained music databases: MusicBrainz, an open music encyclopedia containing data on releases, artists, and labels; Last.fm, a music discovery platform with listening statistics and track-level data; and Discogs, a marketplace-oriented database with detailed release information including genres and country of origin. The 37,255 records overlap partially, as popular releases are contained in all three sources while niche releases may exist in only one. Key challenges include normalizing duration formats and country names, and matching releases as the sources use varying naming conventions.

%% file: sections/03_schema_matching.tex
\section{Schema Matching}
\label{sec:schema_matching}

Schema matching identifies correspondences between attributes in source datasets and the target schema~\cite{rahm2001survey, zhang2023schema}. This section describes how schema matching is handled in the LLM-based and the human pipeline and compares the results of both approaches.

\textbf{Human Pipeline.} As the source schemata are relatively small (5--10 attributes per dataset), the data engineers manually inspected column names and sample values to determine mappings to the target schema. This produced 67 attribute correspondences across all three case studies. For comparison, we also run two non-LLM schema matching methods provided by PyDI: A label-based matcher which compares attribute labels using Monge-Elkan similarity. The best-performing internal similarity metric (Levenshtein or Jaro-Winkler) is automatically selected per use case. An instance-based matcher which tokenizes column values by whitespace, computes TF-IDF vectors, and compares source and target column vectors using cosine similarity. We use the fusion validation set (well-known entities with ground truth values, see Section~\ref{sec:data_fusion}) as the target column representation.

\textbf{LLM Pipeline.} \label{sec:sm_approach}
The LLM pipeline directly prompts GPT-5.2 to perform schema matching. Each source dataset is matched to the target schema using a single LLM call. The prompt consists of a system message instructing the model to act as a schema alignment expert, followed by the source table representation and the target schema. The source representation includes column names, a value summary (unique counts and example values per column), and sample rows rendered as a markdown table. Rows are selected by completeness to maximize informative content. The target schema is provided as a JSON Schema document specifying attribute names, data types, and attribute descriptions. GPT-5.2 returns a JSON array of correspondences, pairing each source column with its target column or indicating no match. For cases where column names are non-descriptive (e.g., ``Attribute\_1'' in the FullContact and MusicBrainz datasets), the prompt explicitly instructs the model to infer attribute semantics from the data values.
The complete prompt is found in the accompanying repository.\footnote{\url{https://github.com/wbsg-uni-mannheim/automatic-data-integration/tree/main/prompts}\label{fn:prompts}}

\input{tables/schema_matching_results}

\textbf{Results.} \label{sec:sm_results}
Table~\ref{tab:sm_results} reports schema matching F1 scores macro-averaged across source datasets for all four approaches.

Both the human data engineers and GPT-5.2 correctly identified all 67 attribute correspondences, achieving perfect F1 scores. For the LLM, this includes two datasets with non-descriptive headers (``Attribute\_1'', ``Attribute\_2'', etc.) where the model differentiated columns based solely on value patterns, for instance distinguishing assets from revenue by value magnitudes. The label-based matcher achieved high precision but low recall (average F1=0.32), correctly matching obvious name pairs (e.g., ``tracks\_track-name'' to ``tracks'') but missing semantically equivalent attributes with different names (e.g., ``imprint'' to ``label''). The instance-based matcher performs worse (average F1=0.19) and fails completely on Games (F1=0.00) as data values do not overlap sufficiently between source and target.

\textbf{Discussion.}\label{sec:sm_limitations} GPT-5.2 was able to discover the same schema correspondences that were manually determined by the data engineers. A major limitation of the applied matching technique is that it assumes single tables and does not cover scenarios where entity data is distributed over several normalized tables. It also does not cover more complex correspondences (1:n or higher-order) as well as data aggregation. If required, approaches to further improve schema matching could include applying duplicate-based schema matching~\cite{naumann2006dataFusionThreeSteps}, as LLMs are quite capable of identifying entity correspondences between datasets (see Section~\ref{sec:em_approach}), or using more advanced LLM-based schema matching techniques~\cite{liu2025magneto}.

%% file: tables/schema_matching_results.tex

\begin{table}[t]
\caption{Schema matching F1 scores (macro-averaged across datasets).}
\label{tab:sm_results}
\centering
\footnotesize
\begin{tabular}{@{}lcccc@{}}
\toprule
\textbf{Case Study} & \textbf{Human} & \textbf{LLM} & \textbf{Label-based} & \textbf{Instance-based} \\
\midrule
Games & 1.00 & 1.00 & .23 & .00 \\
Companies & 1.00 & 1.00 & .35 & .32 \\
Music & 1.00 & 1.00 & .38 & .24 \\
\midrule
\textit{Average} & 1.00 & 1.00 & .32 & .19 \\
\bottomrule
\end{tabular}
\end{table}

%% file: sections/04_normalization.tex
\section{Value Normalization}
\label{sec:normalization}
After schema matching, value normalization translates source data into the formats specified by the target schema~\cite{rahm2000dataCleaning, narayan2022can}. This involves both standard type conversions (dates, numeric scales, country codes) and semantic mappings for categorical attributes where the same concept appears under different names across sources, or taxonomic attributes where sub-concepts need to be mapped to super-concepts in the target schema.

\textbf{Human Pipeline.} \label{sec:norm_approach}
The data engineers manually selected and configured PyDI normalization functions for each source column. For standard data types such as dates, numeric values, and scale modifiers (e.g., ``million,'' ``MEUR''), they specified the expected input format and target output type for PyDI's deterministic normalizers. For categorical attributes, normalization effort varied across use cases. In Games, the data engineers created a manual platform name mapping covering approximately 20 canonical forms with their common aliases (e.g., ``PS4'' $\to$ ``PlayStation~4'', ``SNES'' $\to$ ``Super Nintendo''). In Companies and Music, no manual mapping was performed for taxonomic attributes such as industry or genre.

\textbf{LLM Pipeline.}
The LLM pipeline uses the same code-based normalizers for standard data types. The normalizers are selected automatically through PyDI's column profiling module which detects the semantic type of each column and applies the appropriate normalizer. These normalizers rely on established Python libraries: Babel for locale-aware numeric parsing (handling varying decimal and thousands separators), Pint for unit of measurement conversion, phonenumbers for phone number standardization, and pycountry for country and currency code normalization. For categorical attributes, the LLM pipeline uses GPT-5.2 to map source values to the target value sets defined in the target schemata. To generate the mappings, the LLM uses its background knowledge to resolve synonyms, abbreviations, and hierarchical relationships (see Footnote~\ref{fn:prompts} for the complete prompt). In the Games case study, the LLM maps platform names, genres, and ESRB ratings across all three sources. In Companies, it maps industry labels to the GICS taxonomy. In Music, it maps genre labels to level two concepts in a genre taxonomy. This also eases downstream tasks, as entity matching and data fusion benefit from platform names like ``PS4'' and ``PlayStation~4'' being normalized.

\input{tables/normalization_stats}

\textbf{Results.}
\label{sec:norm_results}
Table~\ref{tab:normalization_stats} presents normalization statistics for the LLM pipeline; since the human pipeline performed only limited normalization, we report only LLM results.
Code-based normalizers handle approximately twice the volume of LLM-based taxonomy mapping (236,610 vs.\ 114,420 values normalized), with high coverage in both categories (96\% for code-based columns, 92\% for LLM-targeted columns). The Games case study dominates LLM usage (4 columns, 84,857 values) due to multiple taxonomy-based attributes, while Music and Companies each require taxonomy mapping for only one column.

\textbf{Discussion.}
\label{sec:norm_limitations}
Compared to the human pipeline, the LLM pipeline normalizes substantially more categorical and taxonomic attributes. Where the data engineers created a small set of platform alias mappings for one use case, the LLM generated comprehensive taxonomy mappings across all categorical columns in all three use cases. Manual inspection of the generated mappings confirmed that they are meaningful and consistent with the target taxonomies. This difference is most pronounced for attributes with many unique values: manually mapping hundreds of genre labels or platform variants is costly, while the LLM handles this at marginal cost. The human time estimates in Section~\ref{sec:runtime_analysis} should therefore be considered a lower bound, as they do not account for the effort that comprehensive taxonomy mapping would have required.

Both normalization approaches have characteristic failure modes. LLM taxonomy mapping retains the original value when no suitable target taxonomy entry exists, leaving unmapped values that propagate to downstream steps. Code-based normalization fails on unknown value formats and lacks semantic understanding: for instance, a duration column containing the value ``2000'' is parsed as-is, even though an album duration of 2,000 minutes is implausible and likely indicates that the scale of the value is seconds. Both limitations could be addressed by coding agents that dynamically generate normalization code informed by background knowledge about the entities being integrated.

%% file: tables/normalization_stats.tex

\begin{table}[t]
\caption{Value normalization statistics by method and case study.}
\label{tab:normalization_stats}
\centering
\scriptsize
\setlength{\tabcolsep}{4pt}
\begin{tabular}{@{}ll r rr r rr@{}}
\toprule
 & & \multicolumn{3}{c}{\textbf{LLM Mappings}} & \multicolumn{3}{c}{\textbf{Code-based}} \\
\cmidrule(lr){3-5} \cmidrule(lr){6-8}
 & \textbf{Source} & \textbf{Col.} & \textbf{Norm.} & \textbf{Total} & \textbf{Col.} & \textbf{Norm.} & \textbf{Total} \\
\midrule
\textit{Games} & DBpedia & 1 & 38,268 & 46,170 & 2 & 83,439 & 91,341 \\
 & Metacritic & 2 & 38,712 & 38,712 & 3 & 59,204 & 59,206 \\
 & Sales & 1 & 7,877 & 7,877 & 3 & 23,631 & 23,631 \\
\midrule
\textit{Companies} & DBpedia & 1 & 6,395 & 6,878 & 2 & 16,429 & 16,963 \\
 & Forbes & 1 & 1,895 & 1,957 & 1 & 1,895 & 1,957 \\
 & FullContact & 0 & 0 & 0 & 1 & 1,053 & 1,056 \\
\midrule
\textit{Music} & Discogs & 1 & 21,273 & 22,627 & 2 & 41,666 & 43,020 \\
 & Last.fm & 0 & 0 & 0 & 1 & 4,635 & 4,635 \\
 & MusicBrainz & 0 & 0 & 0 & 2 & 4,658 & 4,712 \\
\midrule
\multicolumn{2}{@{}l}{\textbf{Total}} & \textbf{7} & \textbf{114,420} & \textbf{124,221} & \textbf{17} & \textbf{236,610} & \textbf{246,521} \\
\bottomrule
\end{tabular}
\end{table}

%% file: sections/05_entity_matching.tex
\section{Entity Matching}
\label{sec:entity_matching}
Entity matching identifies records across different datasets that describe the same real-world entity~\cite{christen2012data, peeters2024entity}. This section compares the data engineers' manually configured entity matchers against ML-based matchers trained on human-labeled and LLM-labeled data.

\textbf{Human Pipeline.}\label{sec:em_human} The data engineers manually configured blocking methods, attribute comparators, and matching thresholds for each dataset pair using PyDI. Blocking keys are derived from entity names (e.g., longest or first token) to prune the comparison space. PyDI provides typed comparators (string, date, numeric) that the data engineers selected for each attribute. Depending on the dataset pair, the data engineers used either rule-based matchers employing attribute weights and a similarity threshold or ML classifiers trained on human-labeled pairs.

\textbf{LLM Pipeline.}
\label{sec:em_approach}
The LLM pipeline replaces both human labeling and manual matcher configuration with LLM-generated training data and automated model selection. We employ an active learning workflow to select entity pairs to be labeled by the LLM. We use embedding-based blocking to generate a pool of candidate pairs: records are embedded using OpenAI's text-embedding-3-small model and the $k{=}20$ nearest neighbors from each other dataset are retrieved. To seed the active learning process, GPT-5.2 labels pairs from this pool in the order of decreasing similarity, stopping for each query entity once at least one positive and two negative matches have been found. Two candidates from the bottom of the similarity ranking are additionally labeled per query to increase diversity. The target number of seed pairs is configurable; the process stops once this target is reached or a labeling budget is exhausted, yielding approximately 100 labeled seed pairs per dataset pair in our setup (see footnote~\ref{fn:prompts} for all prompts).

The seed pairs are used to initialize an active learning loop. Multiple classifiers (RandomForest, XGBoost, GradientBoosting, HistGradientBoosting, LogisticRegression) are trained on string similarity features (Jaccard, Jaro-Winkler, Levenshtein, cosine, and others) computed for each candidate pair, using the same datatype-dependent PyDI similarity metrics as the human pipeline. Hyperparameters are tuned via randomized search. The active learning loop identifies the pairs in the candidate pool on which the classifiers disagree most, measured by the variance of their prediction scores, and selects 100 such candidates for GPT-5.2 to label per iteration. The newly labeled pairs are added to the training set, and all classifiers are retrained. This repeats until the training set reaches a target size or the labeling budget is exhausted. The final training set is augmented with 20\% randomly sampled pairs from the candidate pool to counteract the selection bias toward difficult pairs. The best combination of training set variant, classifier, and threshold is selected per dataset pair by F1 on a GPT-5.2-labeled validation set. Across all dataset pairs, RandomForest and XGBoost are the classifiers most frequently selected.

\input{tables/entity_matching_results}

\textbf{Results.}
\label{sec:em_results}
Table~\ref{tab:em_results} compares three configurations on held-out test sets ranging from 139 to 1,000 manually labeled pairs per dataset pair: The data engineers' manually configured pipeline (Human Config), ML matchers trained on human-labeled data (Human Labels), and the same ML matchers trained on LLM-labeled data (LLM Labels). The Human Labels and LLM Labels columns isolate the effect of label quality, as both use identical feature generation and model selection.

The LLM-labeled matchers achieve the highest average F1 of .937, followed by human-labeled matchers at .916 and the manually configured pipeline at .894. The manually configured matchers are competitive in Games (F1 .930 and .927) and reach the highest individual score in Music Last.fm--MusicBrainz (.977), but perform substantially worse on Music Discogs--MusicBrainz (.800) and Companies (.857 and .870). 

Comparing Human Labels and LLM Labels, performance is nearly identical in Music (F1 $>$ .96 for both) and Companies (within 1.5 percentage points). The largest gain from LLM labeling appears in Games DBpedia--Sales (+14.0\%), possibly due to labeling errors in the manually created training set or because active learning selected more relevant examples compared to the data engineers' selection. In the Games domain, the manually configured baseline likely benefits from its hand-crafted blocking setup, which changes the candidate pool seen by the matcher; therefore, differences between Human Labels and Human Config do not only reflect label quality but also upstream candidate generation effects. The active learning variant achieved the best F1 in five of six cases.

\textbf{Cluster Post-Processing.} False positive correspondences can link unrelated entities into oversized clusters containing more records than the number of source datasets. In order to split such clusters, we remove potentially wrong correspondences by applying maximum bipartite filtering. After cleaning, clusters contain at most three records (one per source), dominated by pairs (62--79\%).

\textbf{Discussion.}
\label{sec:em_limitations}
The results show that ML matchers trained on LLM-generated labels match or exceed the effectiveness of both manually configured matchers and matchers trained on human-labeled data. The effectiveness of the LLM-generated training sets likely results from both the active learning workflow selecting informative pairs and GPT-5.2's ability to correctly label them. Embedding-based blocking with $k{=}20$ neighbors generates a large candidate space (up to 837k pairs across all dataset pairs). Labeling all candidates directly with the LLM would cost approximately \$466, whereas the active learning approach labels only 600--2,658 examples per dataset pair at an overall cost of \$4.95 for all three case studies (Table~\ref{tab:pipeline_costs}), making the active learning-based approach two orders of magnitude cheaper. 

%% file: tables/entity_matching_results.tex

\begin{table}[t]
\caption{Entity matching F1 scores on held-out test sets. \emph{Human Config}: the data engineers' pipeline. \emph{Human Labels} and \emph{LLM Labels}: identical ML matchers trained on human-labeled vs.\ LLM-labeled training data.}
\label{tab:em_results}
\centering
\footnotesize
\begin{tabular}{@{}ll ccc@{}}
\toprule
 & & \textbf{Human} & \textbf{Human} & \textbf{LLM} \\
\textbf{Use Case} & \textbf{Dataset Pair} & \textbf{Config} & \textbf{Labels} & \textbf{Labels} \\
\midrule
\multirow{2}{*}{Games}
 & DBpedia--Metacritic & .930 & .826 & .849 \\
 & DBpedia--Sales & .927 & .839 & .979 \\
\midrule
\multirow{2}{*}{Companies}
 & DBpedia--Forbes & .857 & .954 & .939 \\
 & Forbes--FullContact & .870 & .898 & .897 \\
\midrule
\multirow{2}{*}{Music}
 & Discogs--MusicBrainz & .800 & .991 & .990 \\
 & Last.fm--MusicBrainz & .977 & .988 & .968 \\
\midrule
\multicolumn{2}{@{}l}{\textit{Average}} & .894 & .916 & .937 \\
\bottomrule
\end{tabular}
\end{table}

%% file: sections/06_data_fusion.tex
\section{Data Fusion}
\label{sec:data_fusion}

Data fusion merges all records that describe the same real-world entity into a single consolidated record. If attribute values from different records conflict with each other, a conflict resolution method is applied to resolve the conflict by selecting or generating a single value~\cite{bleiholder2009data, naumann2006dataFusionThreeSteps}. 

\textbf{Conflict Resolution.}\label{sec:fusion_approach} 
The PyDI data integration framework offers a range of different conflict resolution methods including voting, numeric aggregations (e.g., average, median), string-based methods (e.g., longest string, shortest string), temporal (e.g., most recent), and source-aware methods (e.g., source priority based on estimated source accuracy). To perform data fusion, one of these conflict resolution methods is selected for each attribute in the target schema.

\textbf{Human Pipeline.} Graduate students first created validation sets of 10--23 entities per use case. To select entities, they browsed entity clusters containing conflicting attribute values and chose entities whose correct values could be reliably verified through external sources. For each selected entity and each conflicting attribute, they performed web searches to determine the ground truth value, consulting authoritative sources such as official company websites, music databases, or game registries. They then assigned a conflict resolution method to each attribute by choosing from PyDI's library of resolvers. For each attribute, the students considered the data type and domain semantics to select an appropriate method, for instance median for numeric scores and revenue. They iteratively tested different conflict resolution methods by comparing fused values to the corresponding values in the validation set and chose the configuration resulting in the highest value overlap.

\textbf{LLM Pipeline.} The LLM pipeline also generates fusion validation sets that are used to choose conflict resolution methods. The pipeline generated fusion validation sets in two steps: First, GPT-5.2 is prompted to select well-known entities from a random sample of 100 entity groups, i.e., entities for which it is likely easy to gather ground truth (e.g., The Beatles in Music, Apple in Companies). GPT-5.2 selects approximately 30 groups per use case. Second, the model is prompted to generate the value that it perceives as correct for each conflicting attribute. We test two variants: LLM-only validation, where the model generates values solely from its parametric knowledge, and RAG validation, where GPT-5.2 is prompted for each entity and each conflicting attribute to independently verify the correct value by searching the web via the OpenAI web\_search tool. The model can invoke this tool multiple times per call, allowing for query refinement and retrieval from multiple sources.

Using these validation sets, the pipeline evaluates multiple candidate configurations generated by combining heuristic and LLM-based rule selection strategies with iterative refinement, and selects the configuration with the highest validation accuracy (see footnote~\ref{fn:prompts} for all prompts).

\textbf{Results.}\label{sec:fusion_results} Table~\ref{tab:fusion_results} compares fusion test accuracy on held-out test sets of 15--23 entities. The first column shows the accuracy of the students' manually configured fusion strategies (Human Config). The remaining columns show the accuracy of the best pipeline-generated configuration selected using human-created, LLM-only, or RAG validation sets.

\input{tables/fusion_results}

The pipeline with human validation sets outperforms the manual human configuration on average (.816 vs.\ .800), indicating that automated candidate generation finds better resolver assignments even when using the same validation data. For Games, all three automated variants select the same configuration (.866), surpassing the manual configuration (.832). For Companies, the human validation set leads to the highest accuracy overall (.861), while LLM-only achieves .709 and RAG achieves .744. This gap is largely attributable to temporal mismatches: time-sensitive attributes such as headquarters city and revenue are difficult for the LLM to validate, as the source data reflects a specific point in the past while the LLM returns current values. For Music, LLM-only and RAG validation both select the same configuration (.708), close to the human validation set (.721), while the manual configuration achieves .766. On average, LLM-only validation achieves 76.1\% compared to 81.6\% for human validation. RAG validation provides a modest improvement to 77.3\%, suggesting that GPT-5.2's parametric knowledge is largely sufficient for well-known entities. Because the held-out fusion test sets are relatively small (15--23 entities per use case), these differences should be interpreted as indicative. Larger test sets are required to robustly assess small performance gaps.

\textbf{Discussion.}\label{sec:fusion_discussion}
The results show that LLM-generated validation sets can guide fusion configuration selection comparably to human-created ones for domains with predominantly static attributes (Music, Games). The accuracy gap in Companies is largely attributable to temporal mismatches rather than fundamental limitations of the approach. Providing the LLM with metadata about the temporal context of each source could address this.

The selection of well-known entities for validation introduces a bias, as fusion strategies optimized for famous entities may not generalize to less prominent ones. The RAG variant is therefore most relevant not for the use cases studied here, but for enterprise scenarios involving internal or proprietary data sources where entities may not appear in the LLM's training data. In such settings, grounding validation in web or intranet search could substantially improve validation set quality.

%% file: tables/fusion_results.tex

\begin{table}[t]
\caption{Data fusion test accuracy by validation set origin.}
\label{tab:fusion_results}
\centering
\setlength{\tabcolsep}{4pt}
\begin{tabular}{@{}lcccc@{}}
\toprule
 & \textbf{Human} & \textbf{Human} & \textbf{LLM} & \textbf{LLM} \\
\textbf{Use Case} & \textbf{Config} & \textbf{Val.\ Set} & \textbf{Val.\ Set} & \textbf{+ RAG} \\
\midrule
Games     & .832 & .866 & .866 & .866 \\
Companies & .803 & .861 & .709 & .744 \\
Music     & .766 & .721 & .708 & .708 \\
\midrule
\textit{Average} & .800 & .816 & .761 & .773 \\
\bottomrule
\end{tabular}
\end{table}

%% file: sections/07_end_to_end.tex
\section{End-to-End Evaluation}
\label{sec:end_to_end}

The per-step evaluations in the preceding sections rely on manually curated test sets for schema matching, entity matching, and data fusion. While informative, the test sets only cover small subsets of the output data. We therefore complement the step-wise evaluation by reporting end-to-end metrics that compare the size and density of the output datasets to the source datasets.

\textbf{End-to-end Metrics.}\label{sec:e2e_metrics} 
We calculate the following structural end-to-end metrics: Row gain quantifies the size increase of the output dataset by comparing the number of rows of the output dataset to the largest input dataset.
The fusion ratio measures the percentage of output records that result from fusing multiple source records (non-singletons clusters), indicating how much cross-source integration actually took place.  Density change measures how data completeness shifts during integration, reflecting whether fusion fills missing values from complementary sources or introduces sparsity through schema expansion. Its interpretation depends on the integration scenario: with high schema overlap, a density decrease typically indicates quality loss, whereas with low schema overlap a moderate density decrease can be expected and is acceptable if row gain is substantial and attribute coverage increases.

\textbf{LLM Pipeline.} \label{sec:e2e_results} Table~\ref{tab:e2e_metrics} presents the structural metrics for the LLM pipeline. Across all three case studies, the pipeline achieves substantial row gains (27--41\%) over the largest input dataset, demonstrating effective cross-source coverage expansion. Fusion ratios remain modest (8--14\%), indicating that most output records originate from a single source while a meaningful fraction results from cross-source fusion. 

\input{tables/end_to_end_metrics}

\textbf{LLM versus Human Pipeline.} 
Table~\ref{tab:e2e_comparison} contrasts these results against human-configured pipelines. Both approaches produce comparable output sizes for Games (65,794 vs 65,518) and Music (30,779 vs 30,885), with fusion ratios in a similar range (11--15\%). The largest difference appears in Companies: the human pipeline fuses more aggressively (15.1\% vs 8.1\% fusion ratio) but reduces density ($-$5.2pp), while the LLM pipeline improves it (+5.6pp). Conversely, for Music the human pipeline achieves a higher density gain (+5.8pp vs $-$1.8pp).

\input{tables/end_to_end_comparison}

\textbf{Runtime Analysis.}\label{sec:runtime_analysis} We distinguish pipeline configuration, the time spent on producing all artifacts needed to configure a pipeline (schema mappings, labeled training pairs, fusion validation sets), from pipeline execution time, the time spent on the actual data processing once those artifacts exist. All experiments were run on a MacBook Pro (Apple M4, 24\,GB RAM) which calls the OpenAI API for artifact generation. Table~\ref{tab:runtime} summarizes both times per use case. Execution times are comparable, averaging 88\,s for the LLM pipeline and 77\,s for the human-configured one, as data processing is performed by PyDI in both cases. 

The time required for generating the configuration artifacts differs substantially. The LLM pipeline generates all artifacts fully automatically in approximately 1.9\,h per use case, dominated by active learning iterations that repeatedly query the LLM, embed records, and retrain classifiers. For the human pipeline, we estimate a labeling effort of 30\,sec. per entity matching training pair and one minute per fusion validation entity, and add 2--4\,h per use case for non-labeling configuration work: writing schema mappings, implementing normalization rules, and iterative pipeline debugging. These are lower-bound estimates; across the three use cases, the human baseline labels 5,720 training pairs and 60 fusion validation entities, totaling at least 58~person-hours ($\geq$\,19\,h per use case). The time for labeling the test sets is excluded because both pipelines share the same held-out evaluation data. 

\input{tables/runtime_analysis}

\textbf{Cost Analysis.}\label{sec:cost_analysis} Table~\ref{tab:pipeline_costs} details the LLM usage costs for a single pipeline run. The costs are based on GPT-5.2 pricing as of February 2026\footnote{\url{https://openai.com/api/pricing/}}. The total across all three use cases is approximately \$27, dominated by the creation of the RAG-based fusion validation sets (\$21) using the OpenAI web search API. The human baselines required an estimated 58~person-hours for schema mapping, training data labeling, and fusion configuration (Section~\ref{sec:runtime_analysis}), resulting in much higher costs.

\input{tables/pipeline_costs}

\textbf{Discussion.}\label{sec:e2e_discussion} The structural convergence between LLM and human pipelines in output size and fusion ratios confirms that LLMs can replace the human inputs to a traditional integration pipeline (schema mappings, training labels, validation sets) without fundamentally changing integration outcomes. Density results are mixed: neither approach consistently dominates, as the relative advantage depends on how well the automatically selected fusion strategy matches the attribute structure of each use case. Execution times are comparable between pipelines, since both use the same underlying operators and differ only in their configuration artifacts. The key difference is configuration effort: the LLM pipeline requires approximately 1.9\,h per use case at a cost of roughly \$9, compared to at least 19 person-hours for the human baseline. This 10$\times$ reduction in configuration effort, with comparable integration quality, suggests that LLM-configured pipelines are a practical alternative to manual configuration for standard data integration scenarios.

%% file: tables/end_to_end_metrics.tex

\begin{table}[t]
\caption{End-to-end integration metrics for the three case studies.}
\label{tab:e2e_metrics}
\centering
\begin{tabular}{@{}lrrr@{}}
\toprule
\textbf{Metric} & \textbf{Games} & \textbf{Companies} & \textbf{Music} \\
\midrule
Data Sources & 3 & 3 & 3 \\
Total Input Records & 74,951 & 14,016 & 37,255 \\
Fused Record Groups & 7,235 & 1,031 & 4,178 \\
Output Records & 65,518 & 12,768 & 30,885 \\
Fusion Ratio & 11.0\% & 8.1\% & 13.5\% \\
\midrule
Row Gain vs.\ Largest & +18,938 & +2,683 & +8,258 \\
Row Gain \% & +40.7\% & +26.6\% & +36.5\% \\
\midrule
Avg.\ Input Density & 58.7\% & 52.8\% & 72.6\% \\
Output Density & 63.2\% & 58.5\% & 70.8\% \\
Density Change & +4.5pp & +5.6pp & $-$1.8pp \\
\bottomrule
\end{tabular}
\end{table}

%% file: tables/end_to_end_comparison.tex

\begin{table}[t]
\caption{End-to-end integration metrics comparing human versus LLM pipelines.}
\label{tab:e2e_comparison}
\centering
\scriptsize
\setlength{\tabcolsep}{3pt}
\resizebox{\columnwidth}{!}{%
\begin{tabular}{@{}l cc cc cc cc@{}}
\toprule
 & \multicolumn{2}{c}{\textbf{Output Records}} & \multicolumn{2}{c}{\textbf{Fusion Ratio}} & \multicolumn{2}{c}{\textbf{Row Gain \%}} & \multicolumn{2}{c}{\textbf{Density $\Delta$}} \\
\cmidrule(lr){2-3} \cmidrule(lr){4-5} \cmidrule(lr){6-7} \cmidrule(lr){8-9}
\textbf{Use Case} & Human & LLM & Human & LLM & Human & LLM & Human & LLM \\
\midrule
Games & 65,794 & 65,518 & 11.8 & 11.0 & +41.3 & +40.7 & +2.7 & +4.5 \\
Companies & 11,517 & 12,768 & 15.1 & 8.1 & +14.2 & +26.6 & $-$5.2 & +5.6 \\
Music & 30,779 & 30,885 & 14.6 & 13.5 & +36.0 & +36.5 & +5.8 & $-$1.8 \\
\bottomrule
\end{tabular}}
\end{table}

%% file: tables/runtime_analysis.tex

\begin{table}[t]
\caption{Average runtime per use case. Execution includes normalization, matching, and fusion. Configuration refers to one-time artifact generation. Human configuration times are lower-bound estimates.}
\label{tab:runtime}
\centering
\footnotesize
\begin{tabular}{@{}l rr rr@{}}
\toprule
 & \multicolumn{2}{c}{\textbf{Execution}} & \multicolumn{2}{c}{\textbf{Configuration}} \\
\cmidrule(lr){2-3} \cmidrule(lr){4-5}
\textbf{Use Case} & \textbf{LLM} & \textbf{Human} & \textbf{LLM} & \textbf{Human} \\
\midrule
Games     & 181\,s & 73\,s & 1.9\,h & $\geq$\,18\,h  \\
Companies & 38\,s  & 34\,s & 1.7\,h & $\geq$\,20\,h  \\
Music     & 45\,s  & 124\,s & 2.2\,h & $\geq$\,20\,h  \\
\midrule
\textit{Average} & \textit{88\,s} & \textit{77\,s} & \textit{1.9\,h} & $\geq$\,\textit{19\,h} \\
\bottomrule
\end{tabular}
\end{table}

%% file: tables/pipeline_costs.tex

\begin{table}[t]
\caption{LLM usage costs per use case.}
\label{tab:pipeline_costs}
\centering
\footnotesize
\begin{tabular}{@{}l rrr r@{}}
\toprule
\textbf{Pipeline Step} & \textbf{Games} & \textbf{Companies} & \textbf{Music} & \textbf{Total} \\
\midrule
Schema Matching & \$0.02 & \$0.02 & \$0.02 & \$0.06 \\
Normalization & \$0.13 & \$0.36 & \$0.17 & \$0.66 \\
Training Set Generation & \$1.33 & \$1.96 & \$1.66 & \$4.95 \\
Fusion Validation LLM & \$0.22 & \$0.15 & \$0.23 & \$0.60 \\
Fusion Validation RAG & \$7.46 & \$6.07 & \$7.45 & \$20.98 \\

\midrule
\textbf{Total per Use Case} & \textbf{\$9.16} & \textbf{\$8.56} & \textbf{\$9.53} & \textbf{\$27.25} \\
\bottomrule
\end{tabular}
\end{table}

%% file: sections/08_related_work.tex
\section{Related Work}
\label{sec:related_work}

There is intensive research on using LLMs for specific tasks along the integration pipeline~\cite{freire2025llmDataDiscoveryIntegration}, such as schema matching~\cite{narayan2022can, zhang2023schema,liu2025magneto}, value normalization~\cite{brinkmann2024productAttrNormLLMs}, and entity matching~\cite{peeters2024entity,wang2025matchcompare}. There is much less work investigating how LLMs can be employed to automate complete end-to-end data integration workflows. One such work is KGPipes~\cite{hofer2025kgpipes} which uses LLMs within integration pipelines for the construction of knowledge graphs. In addition to relying on a different underlying data model, KGPipes applies LLMs to fewer tasks within the integration pipeline and uses LLMs directly for entity matching, an approach that is more costly and less scalable than the training-data labeling strategy proposed in this paper.

There is ongoing research on data agents~\cite{zhu2025surveydataagents} which perform data discovery, data preparation, and data analysis. Examples include AutoPrep~\cite{fanAutoPrep2025} which uses a multi-agent approach to prepare data for question answering, and SEED~\cite{maddenCafarellaSEED2023} which generates hybrid data curation pipelines that combine LLM querying with cost-effective alternatives such as vector-based caching and LLM-generated code. Similar to our work, SEED employs LLMs to label training data for entity matching, but does not use an active learning workflow.
Recent work on specifying the target outcome of data preparation workflows includes FlowETL~\cite{diprofioFlowETL2025}, which explores example-based specification, and DP-Bench~\cite{chowdhuryDP-Bench2025}, which presents a benchmark for evaluating the automatic creation of ETL workflows given textual descriptions of the target output. 
Recent benchmarks for evaluating the capabilities of LLM agents to build ELT pipelines include Spider 2.0~\cite{Spider2ICLR2025} and ETL-Bench~\cite{jinETL-Bench2025}. While covering end-to-end data preparation workflows, these benchmarks focus on data access and data transformation, but lack explicit ground truth for entity matching and conflict resolution. In contrast, our evaluation explicitly covers both tasks.
Research on synthesizing data preparation pipelines that pre-dates the widespread adoption of LLMs includes AutoPipeline~\cite{AutoPipelineVLDB2021} which assembles string transformations and table-manipulation operators (e.g., Join, GroupBy, Pivot, etc.) into pipelines using search and reinforcement learning. Complementary to our active learning workflow, HILTS~\cite{Barbosa2026} proposes a human-LLM collaborative labeling framework that combines LLM pseudo-labeling with active learning to reduce annotation effort for downstream classifiers.

%% file: sections/09_conclusion.tex
\section{Conclusion}
\label{sec:conclusion}

This paper investigated to what extent LLMs can replace human involvement in configuring end-to-end data integration pipelines. Across three case studies, we compared an LLM-based pipeline against human-configured baselines.

Overall, the LLM pipeline produced similar results as the human-configured pipelines for all three case studies. This was confirmed by the step-wise as well as the structural end-to-end evaluation.  

LLM-based schema matching achieved perfect accuracy (67 correct out of 67 correspondences), including datasets with non-descriptive column headers where the LLM inferred semantics from values alone. For entity matching, LLM-labeled training data produced matchers of comparable quality to human-labeled data (0.937 vs.\ 0.916 average F1); since both pipelines share the same training code, this comparison isolates label quality directly. For data fusion, RAG validation sets achieve 77.3\% average test accuracy compared to 81.6\% for human-created ones, with the gap concentrated in the companies use case (.744 vs.\ .861) where time-sensitive attributes cause systematic mismatches between outdated values in the datasets and the LLM's current knowledge. For domains with predominantly static attributes (Music, Games), the gap is negligible. Configuration time drops on average from an estimated 19 person-hours to 2 hours of unattended compute at approximately \$9 in LLM usage fees, while execution times remain similar (88\,s vs.\ 77\,s). The labor savings are substantial: For our use cases, the LLM pipeline eliminated the need for manual schema mapping, training data labeling, and fusion validation set creation entirely.

\textbf{Limitations.} Our pipeline assumes flat source tables with 1:1 attribute correspondences and does not cover scenarios where entity data is distributed across normalized tables, nor more complex correspondence types (1:n or higher-order correspondences, aggregation). All three case studies use publicly available datasets that could be contained in GPT-5.2's training data, so the strong results, particularly for schema matching and entity labeling, may partly reflect memorized knowledge rather than generic reasoning. Performance on proprietary or domain-specific data outside the LLM's training distribution remains an open question. The temporal knowledge gap observed in companies fusion points to a related limitation: LLM-generated validation data is most reliable for well-known entities with time-independent attributes, and less so for time-sensitive or obscure records. 

\textbf{Future Work.} 
Promising directions for future work include the agent-based refinement~\cite{zhu2025surveydataagents} of the automatic pipeline, as well as exploring the use of the pipeline’s outputs as a starting point for collaborative human–agent refinement workflows~\cite{santos2025interactiveHarmonization}. For value normalization, LLM-based coding agents could dynamically generate transformation functions that combine the robustness of code-based normalization with the LLM's ability to adapt to heterogeneous value formats. The pipeline currently runs strictly forward: each step consumes the output of the previous one without revisiting earlier decisions. An agentic architecture that feeds back downstream results to earlier steps could improve quality, for instance by detecting oversized entity clusters after matching and selectively relabeling the training pairs that contributed to false correspondences, or by identifying values of diverting format in the value clusters before fusion and using this information to refine the normalization step. Human-agent collaboration offers a middle ground where practitioners review and correct LLM decisions at critical points~\cite{santos2025interactiveHarmonization, Barbosa2026}. Today, the research on end-to-end data integration pipelines is hampered by the lack of challenging benchmarks that cover all steps of the integration process (including entity matching and data fusion) with ground truth. We publish the artifacts of our three case studies as an initial step toward such benchmarks.